\documentclass{article}
\usepackage{aikyam}						

\usepackage{booktabs}						
\usepackage{multirow}						
\usepackage{amsfonts}						
\usepackage{graphicx}						
\usepackage{duckuments}						
\usepackage{tcolorbox}

\usepackage[numbers,compress,sort]{natbib}	

\title[When Graph Tokens Sink: A Mechanistic Analysis of Graph Language Models]{When Graph Tokens Sink: A Mechanistic Analysis of Graph Language Models}

\author[Last et al.]{%
Ding Zhang\\
University of Virginia \\\And
Runtao Zhou\\
University of Virginia \\\And
Wenqing Zheng\\
Capital One \\\And
Rizal Fathony\\
Capital One \\\And
Bayan Bruss\\
Capital One \\\And
Chirag Agarwal\\
University of Virginia
}

\newcommand{\xhdr}[1]{\vspace{0em}\noindent{{\bf #1.}}}
\newcommand{\ie}{\textit{i.e., \xspace}}

\newcommand{\hide}[1]{}

\newcommand{\std}[1]{\scriptsize{$\pm$#1}}

\usepackage{colortbl}

\definecolor{Gray}{gray}{0.9}
\definecolor{LightCyan}{rgb}{0.88,1,1}
\definecolor{darkred}{rgb}{0.8,0.1,0.1}
\definecolor{darkyellow}{rgb}{0.95, 0.68, 0.22}
\definecolor{darkgreen}{rgb}{0.1,0.8,0.1}
\newcolumntype{a}{>{\columncolor{Gray}}c}
\newcolumntype{b}{>{\columncolor{white}}c}

\begin{document}

\maketitle
\begin{abstract}
    \looseness=-1 Graph Language Models (GLMs) have become a promising direction for adapting Large Language Models (LLMs) to graph learning tasks. By transforming graph topology and node information into graph tokens, GLMs allow LLMs to jointly process structured graph inputs and textual instructions. Yet, it remains unclear how LLMs internally interpret these graph tokens and whether graph tokens act as meaningful carriers of graph structure. In this work, we analyze how LLMs process graph information through graph-token behavior in representative GLM architectures.

\looseness=-1\textbf{Findings.} We find that the internal saliency of graph tokens in GLMs is not equivalent to graph information utilization. Graph sink tokens consistently emerge as activation-level outliers: they can be identified by massive activation values along a small set of hidden-state dimensions and are biased toward early graph-token positions. However, this activation-level saliency does not imply that these tokens are the main carriers of graph information. Unlike classical attention sinks in language and vision-language models, graph sink tokens do not necessarily attract the largest attention weights from query tokens. Through pruning, repositioning, and swapping interventions, we show that graph sink tokens are not the most important semantic or structural tokens for downstream prediction.

\looseness=-1\textbf{Implications.} Together, these results suggest that after current GLMs map graph structure into the LLM token space, the resulting graph-token representations do not naturally form a fully usable topology-aware internal representation; instead, they exhibit a decoupling between activation-level saliency and graph-semantic utility. This decoupling points to limitations in existing graph-token construction, placement, and alignment mechanisms. 
    
\end{abstract}

\section{Introduction}
\label{sec:intro}
\looseness=-1 Graph Language Models (GLMs) have emerged as a promising framework for adapting Large Language Models (LLMs) to relational learning tasks~\citep{wang2023can, liu2024can, chen2024exploring}. By transforming graph topology and node features into discrete or soft ``\textit{graph tokens},'' GLMs allow transformers to jointly process structured graph inputs alongside textual instructions. While these architectures show strong empirical performance across standard benchmarks~\citep{jin2024large, ren2024survey}, their central representational assumption remains largely unverified: that inherently non-Euclidean graph topologies can be flattened into a sequential token stream and faithfully parsed by an LLM without triggering architectural pathologies~\citep{tang2024graphgpt}.

\looseness=-1 This representational assumption is directly at odds with known transformer dynamics. Indeed, mechanistic interpretability research has shown that transformers are highly prone to structural pathologies when processing sequential data. For instance, in text~\citep{kovaleva2021bert, wei2022outlier} and vision~\citep{luo2025sink} modalities, models routinely develop `\textit{attention sinks}', \ie specific tokens that accumulate massive internal saliency and attention scores not to convey semantic meaning, but merely to stabilize the network's internal computations. This exposes a profound tension for GLMs. Graph tokens are explicitly injected to transmit vital structural information, yet they are subjected to the exact same transformer dynamics known to hijack tokens for non-semantic, architectural reasons. While initial studies have noted sink-like attention patterns in graph inputs~\citep{guan2025attention}, a much deeper vulnerability remains unresolved: when a graph token becomes internally `loud' within the LLM, is the model actually utilizing it for topological reasoning, or is it merely exploiting the tokenized graph structure to build architectural artifacts?

\looseness=-1\xhdr{Present work} In this work, we study how LLMs process graph tokens through the lens of graph sink behavior by auditing the underlying activation mechanics of two representative GLM designs: node-aligned graph tokens (LLaGA) and encoder-generated graph tokens (TEA-GLM). We demonstrate that in graph modalities, `sink' behavior manifests primarily as activation outliers; consequently, we define \emph{graph sink tokens} by their massive magnitude along specific hidden-state dimensions.
Our mechanistic analysis reveals three key findings: First, graph sink tokens consistently emerge as sparse activation-level outliers and are heavily biased toward early graph-token positions, yet this internal prominence does not reliably translate into attention dominance or downstream utility: graph sink tokens do not necessarily receive the largest query-to-graph attention. Second, targeted intervention experiments show that these highly salient tokens contribute surprisingly little to downstream model performance. Finally, our interpretability analysis demonstrates—for the first time—that graph sink tokens predominantly capture weak, domain-level signals rather than the intended task-specific or topology-aware structural information. Together, these results reveal a severe decoupling between activation-level saliency and graph-semantic utility, suggesting that current GLMs do not automatically form usable topology-aware internal representations after mapping graph structure into the LLM token space.

\section{Graph Sink Tokens Emerge as Activation-Level Outliers}
\label{sec:rq1}

\begin{figure}[t]
    \centering
    \includegraphics[width=\linewidth]{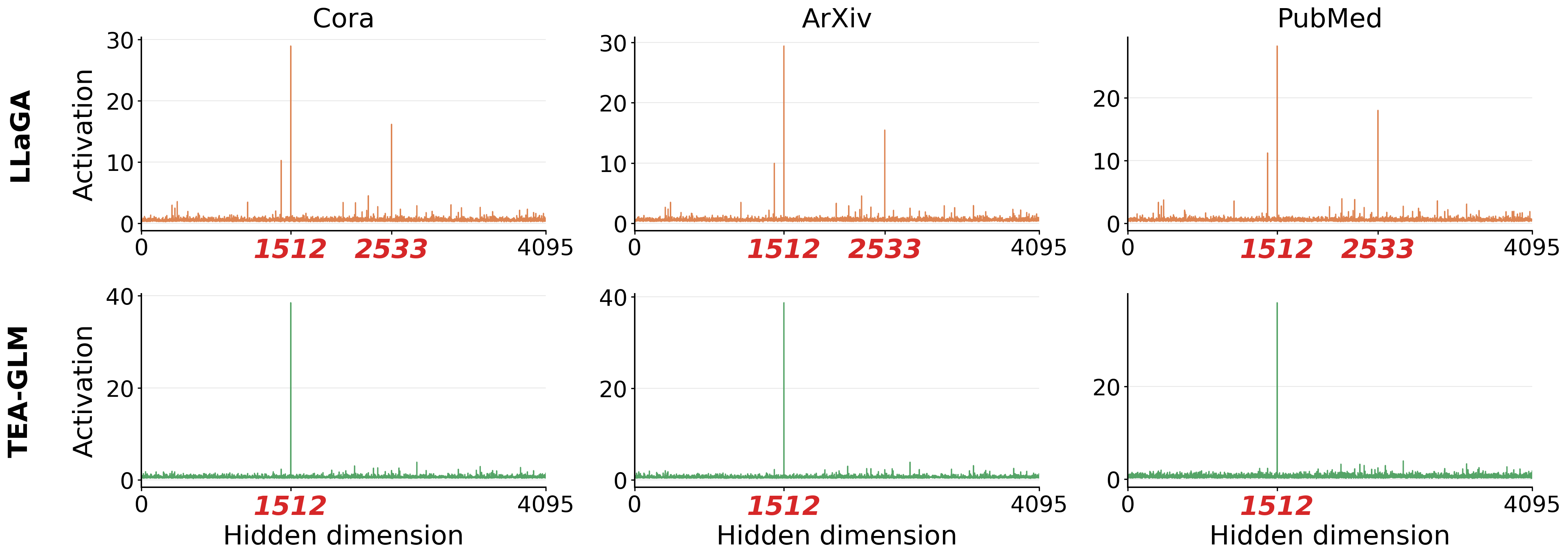}
    \caption{
    \textbf{Activation values across hidden dimensions for detected graph sink tokens on node classification.} We show average activation magnitudes over test samples for LLaGA and TEA-GLM on Cora, Arxiv, and PubMed. Graph sink tokens emerge as sparse activation-level outliers, with large spikes concentrated on a small set of hidden dimensions.
    }
    \label{fig:rq1_activation_nc}
\end{figure}

\begin{figure}[t]
    \centering
    \includegraphics[width=\linewidth]{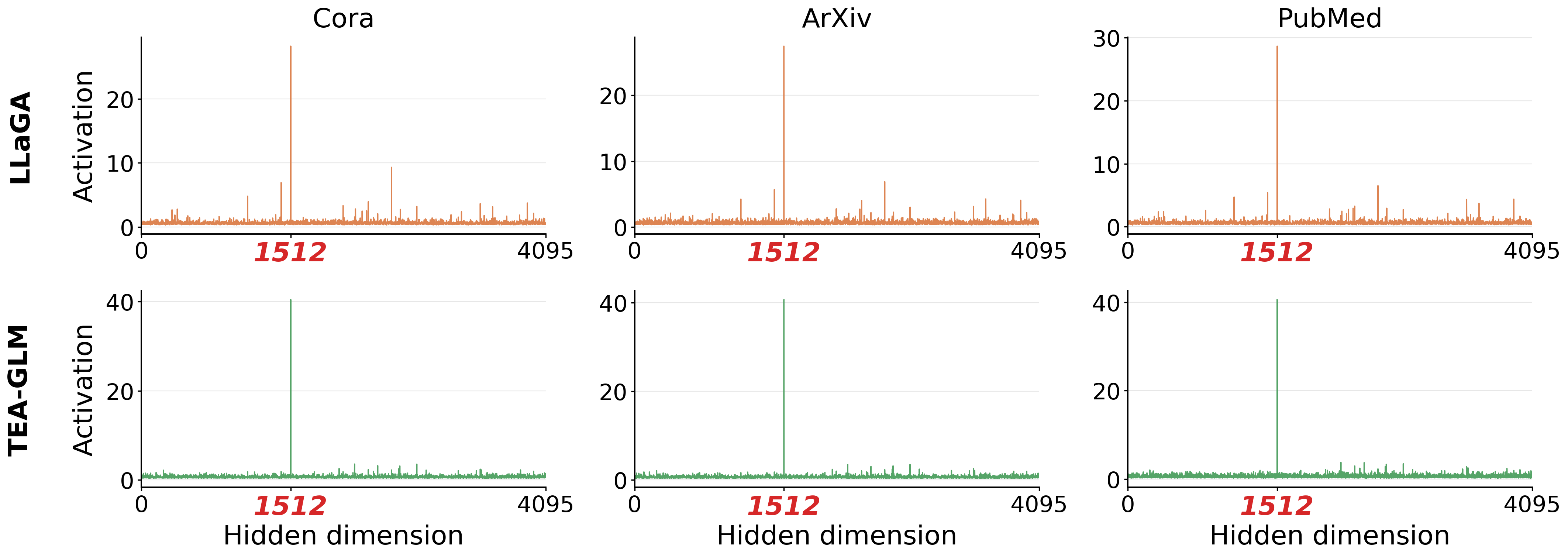}
    \caption{
    \textbf{Activation values across hidden dimensions for detected graph sink tokens on link prediction.}
    The same sparse activation pattern emerges across datasets and architectures, suggesting that the identified sink dimensions are stable across graph learning tasks.
    }
    \label{fig:rq1_activation_lp}
\end{figure}

\xhdr{Graph Sink Definition} Prior works on attention sinks in language and vision-language models show that sink tokens are often linked to massive activation patterns: a small number of tokens produce unusually large values on specific hidden dimensions, and these tokens may later receive disproportionate attention from other tokens \citep{xiao2023smoothquant, darcet2023vision, sun2024massive, kang2025see, luo2025sink}. We call these dimensions \emph{sink dimensions}. In GLMs, we define graph sink tokens from the same activation-based view: a graph token is considered a sink token if its hidden state has a large value on the identified sink dimensions.

\looseness=-1 Following \citep{kang2025see, luo2025sink}, let a GLM take a mixed-token sequence containing both text tokens and graph tokens. We denote the set of graph-token indices by $\mathcal{I}_{g}$, and let $\mathbf{x}^{l}_{j} \in \mathbb{R}^{d}$ be the hidden state of token $j$ at layer $l$, where $d$ is the LLM hidden dimension. The set of graph sink tokens at layer $l$ is defined as:
\begin{equation}
\widehat{\mathcal{I}}^{l}_{g}
=
\left\{
j \in \mathcal{I}_{g}
\ \middle|\
\phi\left(\mathbf{x}^{l-1}_{j}\right) \geq \tau
\right\},
\label{eq:graph_sink_definition}
\end{equation}
where $\phi(\cdot)$ is a sink characteristic function and $\tau$ is a predefined threshold. In this work, we use activation magnitude as the sink characteristic. Given a set of sink dimensions $\mathcal{D}_{\mathrm{sink}}$, we define:
\begin{equation}
\phi\left(\mathbf{x}^{l-1}_{j}\right)
=
\max_{\bar{d} \in \mathcal{D}_{\mathrm{sink}}}
\left|
\mathrm{RMSNorm}\left(\mathbf{x}^{l-1}_{j}\right)_{\bar{d}}
\right|.
\label{eq:sink_characteristic_function}
\end{equation}
Here, $\bar{d}$ denotes one sink dimension, and $\mathrm{RMSNorm}$ denotes Root Mean Square Normalization. Prior work has shown that sink dimensions can be language-model specific \citep{sun2024massive, kang2025see, luo2025sink, sun2026spike}. For example, LLaMA2-7B \citep{touvron2023llama} has been shown to contain sink dimensions such as $\{1415,2533\}$. Since the GLMs studied here are built on LLaMA-family backbones, we ask whether their graph tokens simply inherit these dimensions, or whether graph-token processing induces new sink dimensions.

\looseness=-1\xhdr{GLM Choices} We study two representative GLMs (LLaGA \citep{chen2024llaga} and TEA-GLM \citep{wang2024llms}) as they represent two common ways of passing graph information to an LLM. LLaGA uses a node-aligned graph-token sequence, where we use its Neighborhood Detail (ND) template: the center node is placed at graph sequence index $0$ and its $k$-hop neighborhood is organized into a fixed-shape tree by sampling neighbors at each hop. If the number of sampled neighborhood is smaller than the budget, \texttt{[PAD]} tokens are inserted to maintain graph-token length. TEA-GLM follows a different design. It first uses a GNN encoder to aggregate neighborhood structure into a central-node representation and then projects this representation into a fixed number of learnable graph-token embeddings through a linear projector. These graph tokens are aligned with the LLM hidden space through contrastive learning against principal components of LLM token embeddings. We choose these models for two reasons: i) they represent two broad GLM families: node-aligned graph tokens and encoder-embedded graph tokens; and ii) both models use fixed-length graph-token sequences, which allows us to compare graph sink positions across samples, where $K$ is the number of graph tokens.

\begin{figure}[t]
    \centering
    \includegraphics[width=\linewidth]{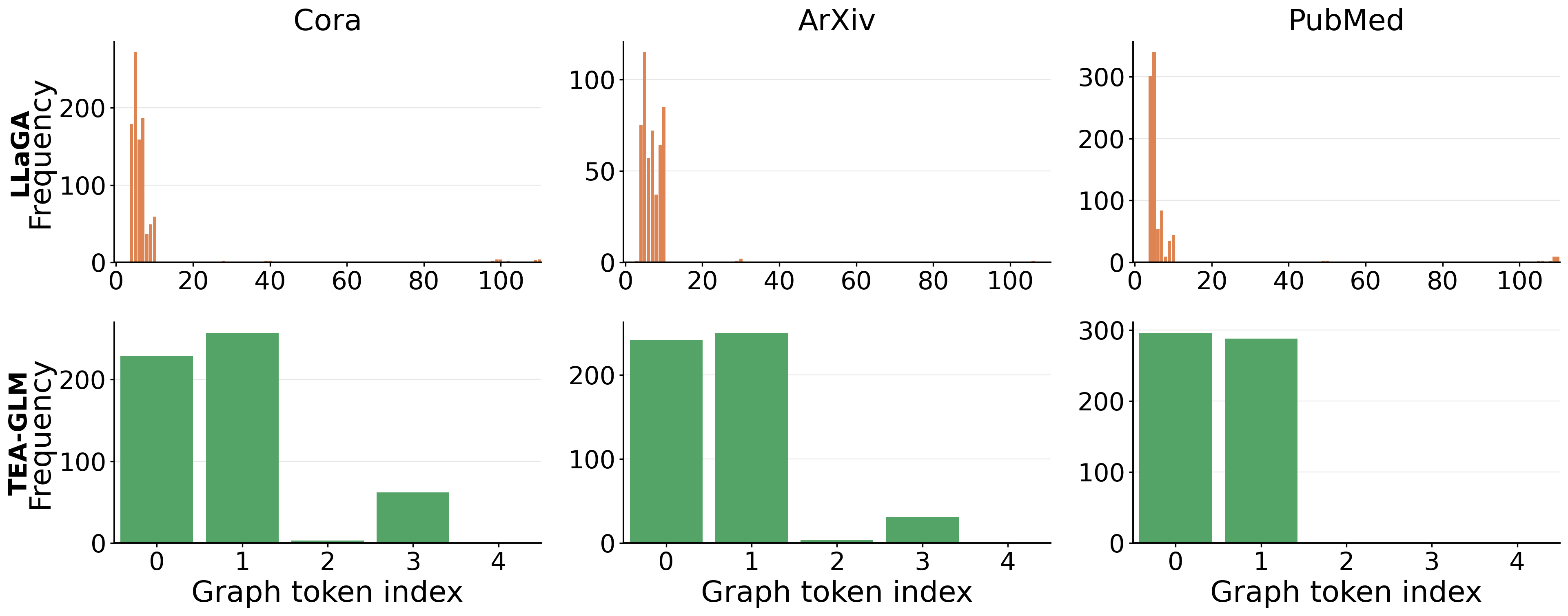}
    \caption{
    \textbf{Distribution of detected graph sink token positions on node classification task.}
    For LLaGA, $K=111$; for TEA-GLM, $K=5$.
    Graph sink tokens are biased toward early graph-token positions.
    In LLaGA, a great portion of detected sink tokens are \texttt{[PAD]} tokens.
    }
    \label{fig:rq1_nc_sink_distribution}
\end{figure}

\begin{figure}[t]
    \centering
    \includegraphics[width=\linewidth]{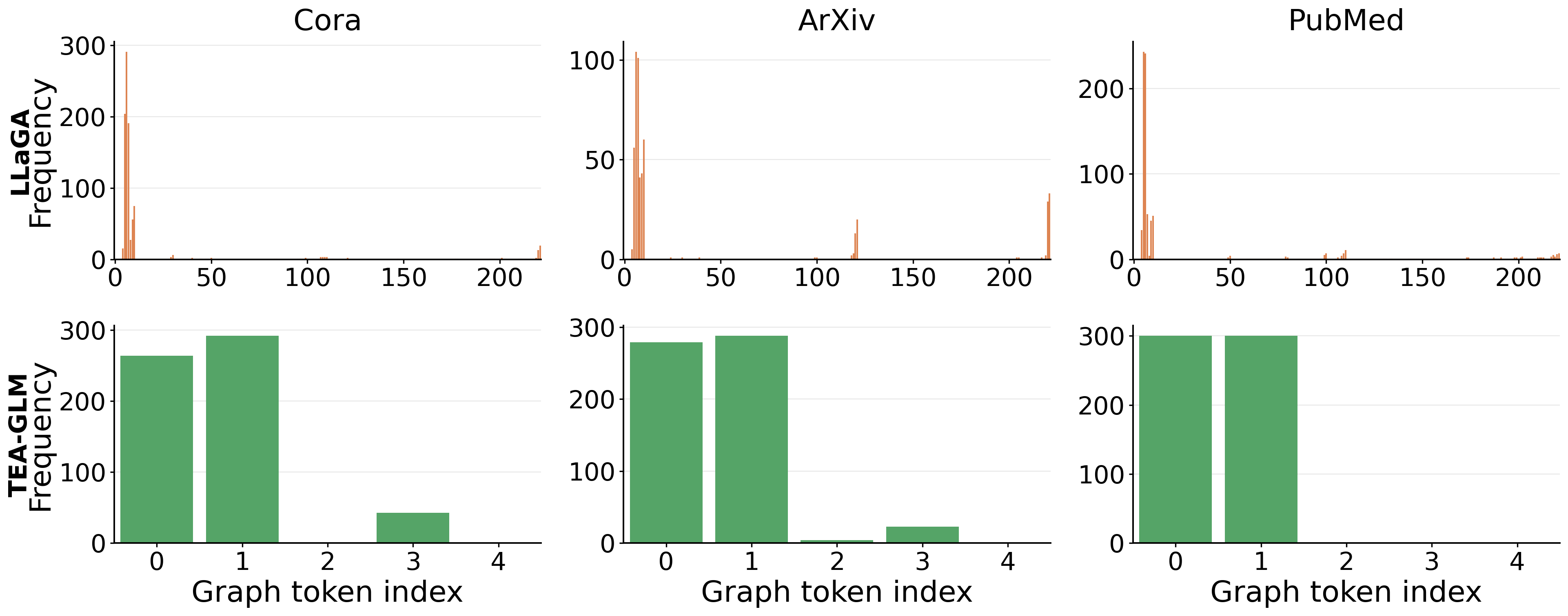}
    \caption{
    \textbf{Distribution of detected graph sink token positions on link prediction task.}
    For LLaGA, $K=222$; for TEA-GLM, $K=5$.
    The position pattern is consistent with node classification: graph sink tokens mainly appear near the beginning of the graph-token sequence.
    }
    \label{fig:rq1_lp_sink_distribution}
\end{figure}

\looseness=-1\xhdr{Sink Activation Magnitudes} We run experiments on Cora \citep{yang2016revisiting}, Arxiv \citep{hu2020open}, and PubMed \citep{sen2008collective}, covering both node classification and link prediction. We set $\tau=15.0$ in all settings for both GLMs. Figures~\ref{fig:rq1_activation_nc} and~\ref{fig:rq1_activation_lp} show the average hidden activation magnitude of detected graph sink tokens on two tasks. For LLaGA, we average over $500$ test samples per dataset; for TEA-GLM, we average over $300$ test samples per dataset. The pattern is sharp and consistent. Across datasets and tasks, most hidden dimensions stay near a low activation range, while only a few dimensions form large spikes. In LLaGA, dimension $1512$ repeatedly appear as dominant sink dimensions for both tasks, and dimension $2533$ is another sink dimension for node classification task. In TEA-GLM, dimension $1512$ is the dominant sink dimension across all datasets and both tasks. This suggests that graph sink behavior is not only inherited from known LLM sink dimensions, such as $2533$ in LLaMA-based models. The repeated appearance of dimension $1512$ across both GLM designs points to sink behavior that emerges after graph information is mapped into the LLM token space.

\looseness=-1\xhdr{Sink Token Distributions} We next ask where these graph sink tokens appear in the graph-token sequence. For LLaGA, the graph-token sequence length is $K=111$ for node classification and $K=222$ for link prediction. For TEA-GLM, the graph-token sequence length is fixed as $K=5$ for both tasks. Figs.~\ref{fig:rq1_nc_sink_distribution}-\ref{fig:rq1_lp_sink_distribution} show the frequency of detected graph sink tokens at each graph-token index. The distribution is not uniform. Across tasks, datasets, and models, graph sink tokens are biased toward early graph-token positions. For TEA-GLM, most sink tokens appear at positions $0$ and $1$, indicating that the first few encoder-generated graph-token slots are most likely to become activation outliers. LLaGA shows the same early-position bias, but the distribution is less compressed because its graph-token sequence is much longer. LLaGA also lets us inspect whether these sink positions correspond to meaningful graph roles. Under the ND template, index $0$ is the center node, while later positions correspond to sampled neighbors or \texttt{[PAD]} tokens. We find that the top-$2$ graph sink tokens with the largest activation scores are always \texttt{[PAD]} tokens across all datasets and both tasks. By contrast, the center-node token at index $0$ is never identified as a graph sink token. This already suggests a gap between activation-level saliency and graph-semantic importance: the most salient graph tokens by activation are not necessarily the most meaningful graph tokens by structure.

\begin{tcolorbox}
\looseness=-1\textbf{Key Takeaway.} Graph sink tokens consistently emerge as activation-level outliers, identified by large values on a small set of hidden dimensions and biased toward early graph-token positions.
\end{tcolorbox}
\section{Activation-Level Saliency Does Not Imply Attention Dominance}
\label{sec:rq2}

Sec.~\ref{sec:rq1} shows that graph sink tokens emerge as activation-level outliers: they have large responses on a small set of hidden dimensions and are biased toward early graph-token positions. This raises a critical question: 
\textit{do these exceptionally ``loud'' tokens also dictate the routing of information by attracting the majority of attention weights?} 

\begin{figure}[t]
    \centering
    \includegraphics[width=\linewidth]{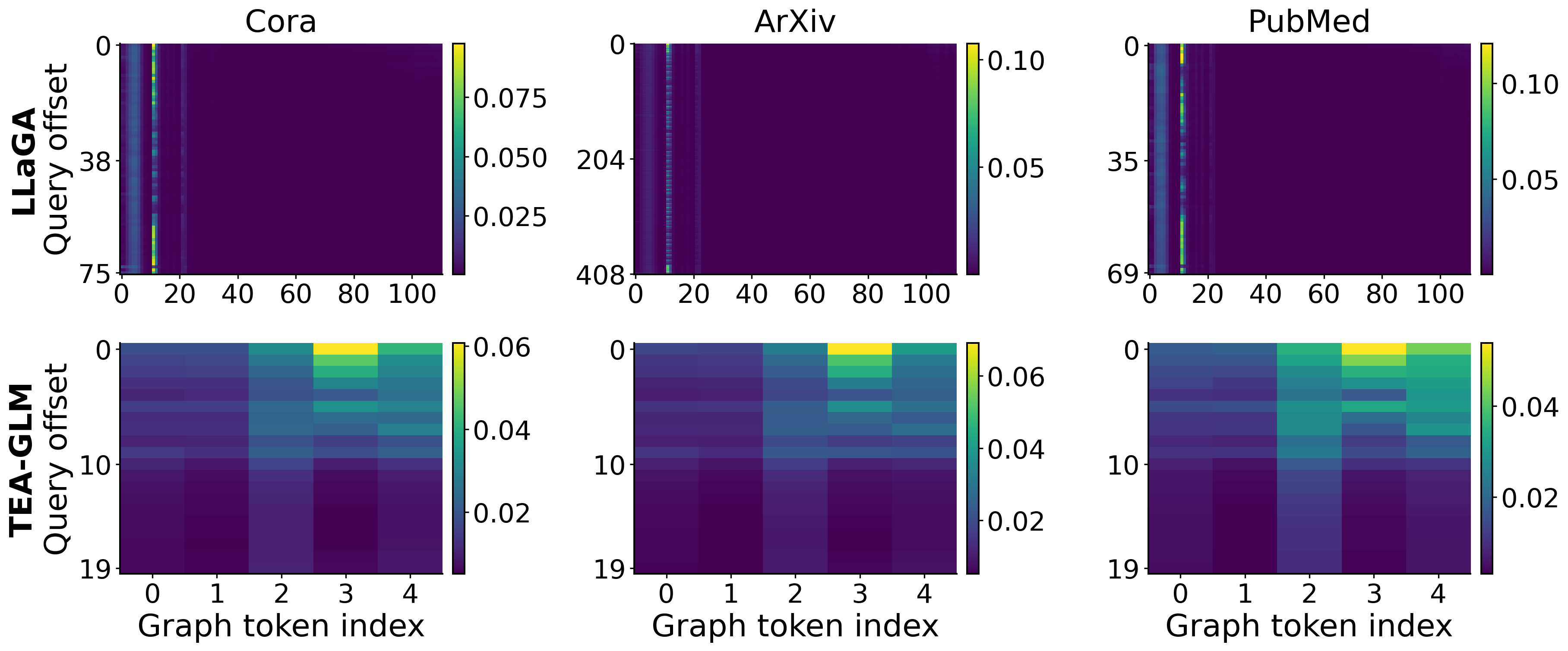}
    \caption{
    \looseness=-1\textbf{Query-to-graph attention maps for node classification for both models across three datasets.}
    Attention weights are averaged over heads and test samples. In LLaGA, several graph-token positions receive stable attention across query offsets, but sink tokens are not necessarily the highest-attended tokens. In TEA-GLM, high-attention regions often appear on later graph-token indices rather than the main sink-token positions \{$0, 1$\}.
    }
    \label{fig:rq2_nc_query_attention}
\end{figure}

\looseness=-1\xhdr{Query-to-Graph Attention} We first analyze attention from query to graph tokens during the pre-fill stage. For each model and dataset, we average attention weights over heads and test samples, and visualize how query-token positions attend to each graph-token index. Fig.~\ref{fig:rq2_nc_query_attention} shows node classification results on Cora, Arxiv, and PubMed (link prediction results are provided in Appendix~\ref{app:rq2_link_attention}). The TEA-GLM results show a clear separation between sink-token and high-attention positions. As shown in Sec.~\ref{sec:rq1}, TEA-GLM graph sink tokens mostly appear at indices $0$ and $1$. However, the average attention is often stronger on later graph tokens, especially indices $2$--$4$. Conversely, the attention maps of LLaGA contain narrow vertical bands over several graph-token positions, indicating that these positions receive stable attention across query offsets. Some of these bands overlap with the graph sink region identified in Sec.~\ref{sec:rq1}. However, this attention is not clearly query-specific, and the average attention on sink positions is not necessarily higher than on non-sink positions. Since the top LLaGA sink tokens with the highest magnitudes are often \texttt{[PAD]} tokens, this stable attention is unlikely to correspond directly to useful node or neighborhood information.

\begin{figure}[t]
    \centering
    \includegraphics[width=\linewidth]{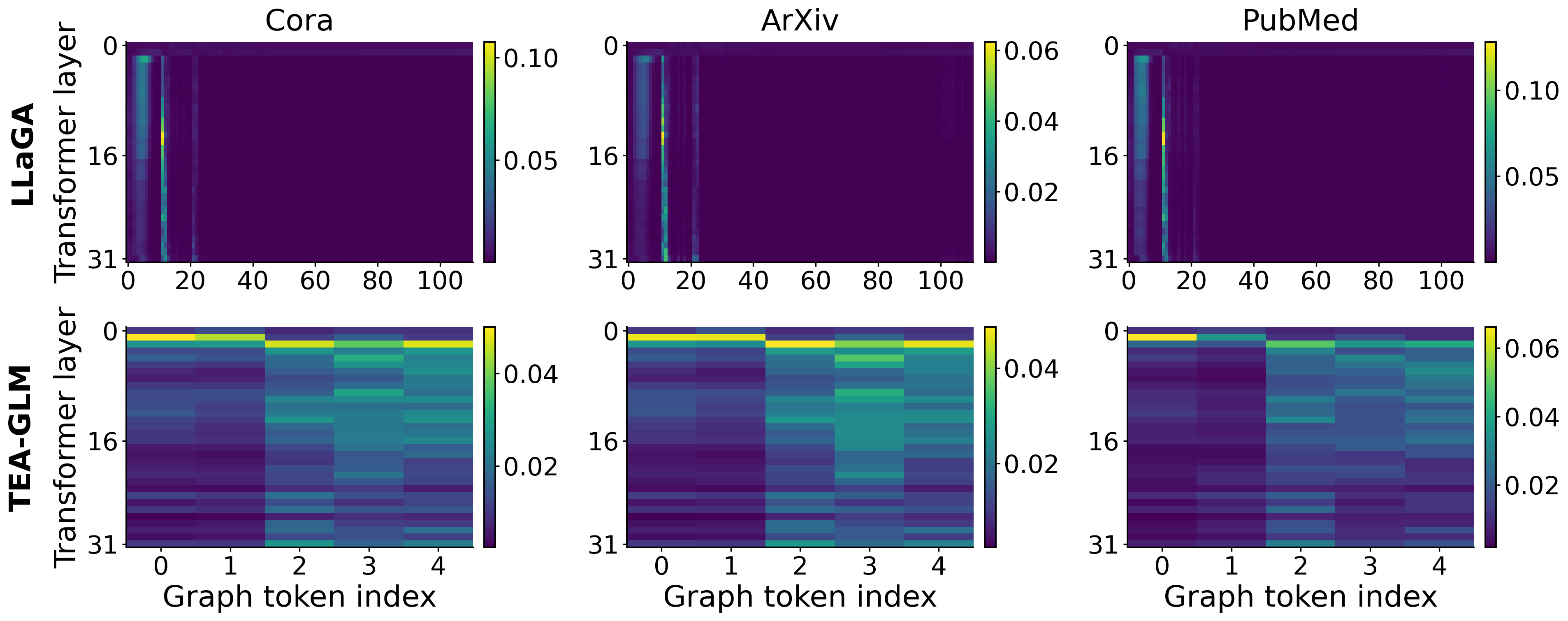}
    \caption{
    \looseness=-1\textbf{Layer-wise query-to-graph attention maps for node classification for both GLMs.}
    Attention weights are averaged over heads and test samples, and the y-axis denotes transformer layers.
    TEA-GLM assigns stronger attention to later graph-token indices than to the main sink-token positions, while LLaGA shows stable vertical attention bands that are not exclusive to graph sink tokens.
    }
    \label{fig:rq2_nc_layer_attention}
\end{figure}

\looseness=-1\xhdr{Layer-wise Attention Patterns} Next, we check whether this separation also appears across transformer layers. Fig.~\ref{fig:rq2_nc_layer_attention} shows graph-token attention as a function of layer. For TEA-GLM, later graph-token indices again receive stronger attention than the main sink-token positions $0$ and $1$, especially in lower and middle layers. This shows that the mismatch between activation saliency and attention dominance is not an artifact of averaging over query positions; it is visible throughout the model. For LLaGA, the layer-wise plots again show thin vertical bands at fixed graph-token positions. These bands persist across layers, suggesting that LLaGA repeatedly routes some attention to the same graph-token positions. Yet the attention is spread across both sink and non-sink regions, rather than being concentrated only on detected graph sink tokens. This evidence supports that LLaGA graph sink tokens can participate in stable attention patterns, but they are not the sole or strongest pathway through which query tokens access graph information. This observation is consistent with recent findings on sink behavior in text-only LLMs.~\citet{sun2026spike} show that activation spikes and attention sinks often co-occur, but they are not strictly coupled. Our results suggest that the same distinction is important for graph tokens: a graph token can be highly salient in the hidden activation space without becoming the dominant target of query-to-graph attention.

\looseness=-1 \xhdr{Summary} Together, the two attention views point to the same conclusion. Graph sink tokens in GLMs are activation-salient, but this saliency does not reliably translate into attention dominance. TEA-GLM provides the clearest case, where sink tokens are located at early graph-token slots but attention often shifts to later graph tokens. LLaGA shows partial overlap between sink positions and stable attention bands, but these sink positions are often \texttt{[PAD]} tokens and are not consistently the highest-attended tokens. Therefore, graph sink tokens should not be treated as direct graph-domain versions of classical attention sinks.

\begin{tcolorbox}
\textbf{Key Takeaway.} Graph sink tokens are activation-salient, but they do not necessarily become the dominant attention targets of query tokens.
\end{tcolorbox}
\section{Graph Sink Tokens Are Not the Main Semantic or Structural Carriers}
\label{sec:rq3}

In Sec.~\ref{sec:rq2}, we show that graph sink tokens are activation-level outliers, yet they do not necessarily receive dominant query-to-graph attention. This raises a functional question: \textit{do these highly activated graph tokens actually carry the graph information used for prediction?} To answer this question, we move from observational analysis to direct graph-token interventions. We test whether removing, swapping, or repositioning graph sink tokens causes meaningful performance degradation.

\looseness=-1 \xhdr{Intervention setup} For each test sample, we first run the original model and identify graph sink token positions using Eq.~\ref{eq:graph_sink_definition}. These positions are then fixed for all intervention runs on the same sample. We consider three interventions. \textsc{Top-$2$} sink removes the two graph sink tokens with the largest activation magnitudes. \textsc{Non-sink} removes two randomly selected non-sink graph tokens, serving as a control for the effect of shortening the graph-token sequence; we average this setting over $15$ random seeds. \textsc{Swap} exchanges two sink-token positions with two randomly selected non-sink positions while keeping all token embeddings in the prompt; we average this setting over five random seeds.

\begin{table}[h]
    \centering
    \setlength{\tabcolsep}{4pt}
    \renewcommand{\arraystretch}{0.95}
    \caption{
    Node classification performance under graph-token interventions.
    We compare the baseline model with pruning the top-$2$ graph sink tokens, randomly pruning two non-sink graph tokens, and swapping sink and non-sink token positions. Non-sink pruning is averaged over 15 random seeds, and swapping is averaged over five random seeds. \textcolor{red}{Red} values denote cases where non-sink pruning causes a larger performance drop than top-$2$ sink pruning and baseline results.
    }
    \label{tab:rq3_pruning_results}
    \begin{tabular}{llccc}
        \toprule
        Model & Intervention & Arxiv & Cora & PubMed \\
        \midrule

        \multirow{4}{*}{LLaGA}
        & Baseline & 77.00 & 88.40 & 94.60 \\
        & Top-$2$ Sink & 77.00 & 88.00 & 95.00 \\
        & Non-sink & \textcolor{red}{74.36}\std{1.44} & \textcolor{red}{80.48}\std{2.47} & \textcolor{red}{89.16}\std{4.58} \\
        & Swap & 76.64\std{0.27} & 87.74\std{0.39} & 94.12\std{0.27} \\
        \midrule

        \multirow{4}{*}{TEA-GLM}
        & Baseline & 56.67 & 13.33 & 83.67 \\
        & Top-$2$ Sink & 56.67 & 13.00 & 83.67 \\
        & Non-sink & \textcolor{red}{44.40}\std{1.13} & \textcolor{red}{12.80}\std{1.72} & {83.40}\std{1.09} \\
        & Swap & 56.40\std{0.43} & 12.87\std{0.18} & 83.27\std{0.43} \\
        \bottomrule
    \end{tabular}
\end{table}

\xhdr{Sink-token interventions have limited impact}
Table~\ref{tab:rq3_pruning_results} shows that pruning the top-$2$ graph sink tokens has little effect on node classification for both LLaGA and TEA-GLM. In contrast, pruning random non-sink tokens leads to larger drops in several settings, especially for LLaGA. We observe a related trend in link prediction, reported in Appendix~\ref{app:intervention}, Table~\ref{tab:app_rq3_link_intervention_results}: although the contrast is clearer in node classification, both tasks show that graph sink tokens are not the main graph-information carriers. Across node classification and link prediction, the most activation-salient graph tokens are often less important than ordinary non-sink tokens for prediction.

In LLaGA, graph tokens are designed to preserve neighborhood structure through the ND template, yet the sink tokens with top-$2$ magnitudes are not the ones whose removal degrades the performance the most. In TEA-GLM, graph tokens are learned as high-dimensional projections of GNN representations, but pruning sink tokens still has little impact. Moreover, on some TEA-GLM datasets, even pruning non-sink tokens causes only small changes, suggesting that the model may not consistently use graph-token semantics in zero-shot prediction.

The swap intervention leads to the same conclusion. Exchanging sink and non-sink positions changes performance only mildly, meaning that sink tokens are not strongly tied to sample-specific graph content or to a uniquely important position. For LLaGA, we further test moving sink tokens to the front of the graph-token sequence. Since LLaGA's ND template encodes graph topology through fixed token positions, such repositioning disrupts the original correspondence between sequence position and neighborhood structure. However, the results in Appendix~\ref{app:intervention}, Table~\ref{tab:rq3_reposition_results}, show little consistent impact on performance, suggesting that these sink tokens are not reliable graph summary or routing tokens. Together, pruning, swapping, and repositioning all suggest that graph sink tokens are not the main carriers of graph-structural information.

\begin{figure}[h]
    \centering
    \includegraphics[width=\linewidth]{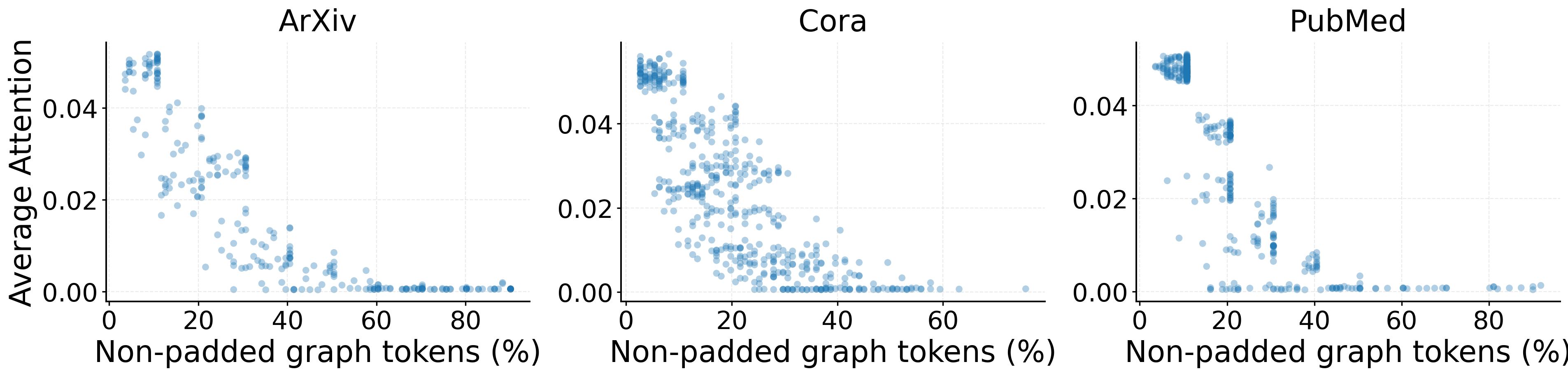}
    \caption{
    \textbf{Relationship between graph-token sparsity and attention to top-$2$ graph sink tokens in LLaGA.}
    The x-axis denotes the percentage of non-padded graph tokens in the graph-token sequence, and the y-axis denotes the average attention assigned to the top-$2$ graph sink tokens.
    Across Arxiv, Cora, and PubMed, attention to the top-$2$ sink tokens decreases as the proportion of non-padded graph tokens increases.
    }
    \label{fig:rq3_llaga_nonpad_sink_attention}
\end{figure}

\xhdr{Sink attention is tied to graph-token sparsity}
The intervention results suggest that LLaGA graph sink tokens are not the main information-bearing tokens. This is consistent with our earlier observation that the strongest LLaGA sink tokens are usually \texttt{[PAD]} tokens. To better understand this behavior, we ask whether attention to these sink tokens depends on how much real graph content appears in the prompt. For each test sample, we compute the percentage of non-padded graph tokens in the graph-token sequence and measure the average query-to-graph attention assigned to the top-$2$ graph sink tokens. Fig.~\ref{fig:rq3_llaga_nonpad_sink_attention} reports this relationship across Arxiv, Cora, and PubMed.

Across all three datasets, we observe a clear decreasing trend: samples with fewer non-padded graph tokens assign higher attention to the top-$2$ sink tokens, while samples with more non-padded graph tokens assign less attention to these sink positions. Since the top-$2$ LLaGA sink tokens are consistently \texttt{[PAD]} tokens, this suggests that sink-token attention is partly tied to prompt sparsity. When the graph-token block contains less real neighborhood information, the model allocates more attention to padded sink positions; when more real graph tokens are present, attention shifts away from these padding-based sinks. This further supports our claim that LLaGA graph sink tokens are not reliable carriers of graph-semantic content, but are strongly shaped by the graph-token construction and padding pattern.

\begin{figure}[h]
    \centering
    \includegraphics[width=\linewidth]{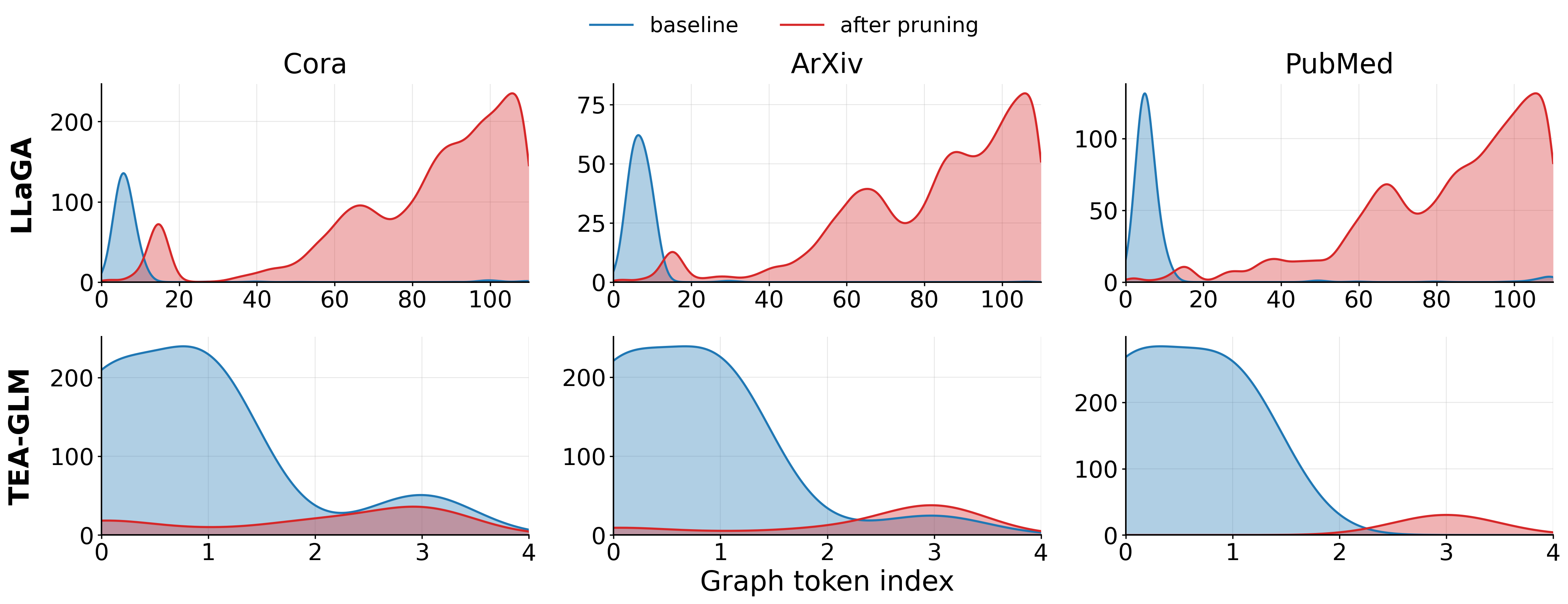}
    \caption{
    \textbf{Sink-position distribution shifts before and after pruning all identified graph sink tokens.} For LLaGA, sink tokens appear again after pruning but become more broadly distributed across graph-token positions. For TEA-GLM, activation magnitudes decrease after pruning, and the remaining graph tokens rarely satisfy the sink criterion. Note that the subplots for TEA-GLM are being smoothed for visualization purposes. 
    }
    \label{fig:rq3_sink_shift}
\end{figure}

{\vspace{0em}\noindent{{\bf Do new sink tokens appear after pruning?}}}
A natural follow-up question is whether removing the original sink tokens simply causes other graph tokens to take over the sink role: if sink behavior is not tied to graph semantics, then it may reappear at other graph-token positions after the original sinks are removed. To test this, after pruning \textit{all} detected sink tokens, we run sink detection again on the pruned graph-token sequence and compare the sink-position distributions before and after pruning in Figure~\ref{fig:rq3_sink_shift}.

For LLaGA, sink tokens appear again after pruning, but their positions become much more spread out across the graph-token sequence rather than remaining concentrated near the earlier biased sink positions. This indicates that LLaGA maintains high-activation behavior even after the original sink tokens are removed, but the sink role is redistributed across many graph-token positions. In other words, pruning does not eliminate the activation-sink pattern in LLaGA; it changes where it occurs and increases the spread of high-activation tokens.

TEA-GLM shows the opposite pattern. After pruning the original sink tokens, the overall activation magnitude decreases, and the remaining graph tokens no longer satisfy the sink criterion in most cases. This suggests that TEA-GLM sink behavior is more localized: once all sink tokens are pruned, the model does not create new sink tokens elsewhere. Combined with the limited performance change under sink pruning, this further suggests that TEA-GLM graph sink tokens are activation-level artifacts rather than reliable carriers of graph semantics.

\looseness=-1\xhdr{Summary} Across pruning, swapping, repositioning, token sparsity analysis, and post-pruning sink detection experiments, graph sink tokens do not behave as the main semantic or structural carriers in graph language models. Removing the top-$2$ magnitude sink tokens usually causes little performance drop, while removing random non-sink tokens can be more harmful, especially for LLaGA. After pruning, LLaGA redistributes sink behavior across many graph-token positions, whereas TEA-GLM largely loses sink-qualified activations. These results connect back to our central observation: activation-level saliency does not imply graph-information utility.

\begin{tcolorbox}
\textbf{Key Takeaway.} Graph sink tokens are not the main graph information carriers.
\end{tcolorbox}
\section{Mechanistic Evidence for Limited Graph-Token Semantics}
\label{sec:rq4}

\looseness=-1 Sec.~\ref{sec:rq3} shows that graph sink tokens are not prediction-critical, \ie pruning them causes little performance change, and post-pruning detection suggests that sink behavior is not tied to a fixed set of semantically essential graph tokens. We now ask a complementary mechanistic question: \textit{what information is exposed by graph-token hidden states inside the LLM?} To answer this, we apply a logit lens analysis \citep{nostalgebraist2020logitlens, geva2021transformer, belrose2023eliciting, agarwal2025rethinking} to TEA-GLM on node classification. We focus on TEA-GLM because it uses a compact fixed graph-token block, $K=5$, making token positions comparable across samples, datasets, and layers. We select samples where graph sink tokens appear at positions g0 and g1, and decode each graph-token hidden state at each transformer layer into the LLM vocabulary space.

\looseness=-1\xhdr{Logit lens setup}
For each dataset, layer, and graph-token position, we aggregate selected test samples and record the vocabulary token that most frequently appears as the top-$1$ decoded token after projecting the graph-token hidden state through the LM head. We also report the average probability of this decoded token across selected samples. Thus, the text inside each cell denotes the most frequent decoded token, while the heatmap color denotes its average probability. This analysis tests whether graph-token hidden states expose task-relevant labels, node semantics, or topology-related concepts.

\begin{figure}[h]
    \centering
    \includegraphics[width=\linewidth]{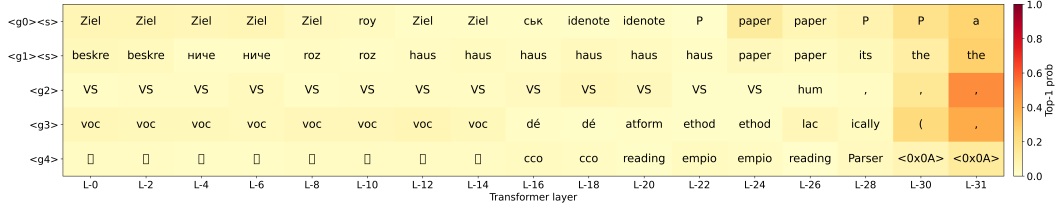}
    \caption{
    \textbf{Logit lens analysis for TEA-GLM graph tokens on PubMed node classification.}
    We select samples where graph sink tokens occur at positions g0 and g1.
    Each cell shows the most frequent top-$1$ decoded vocabulary token at a given graph-token position and transformer layer, while the color indicates its average probability across samples.
    Sink-token positions frequently decode to generic domain terms such as \texttt{paper} in later layers, but the overall probabilities remain low.
    }
    \label{fig:tea_glm_logit_lens}
\end{figure}

\looseness=-1\xhdr{Decoded graph-token states are weakly topology-aware} 
Fig.~\ref{fig:tea_glm_logit_lens} shows limited evidence that decoded graph-token states expose topology-aware or task-specific information. Additional results on Cora and Arxiv are reported in Appendix~\ref{app:logit_lens}. Across layers, many decoded outputs are fragmented subwords, punctuation marks, or generic terms, rather than class labels or structural concepts. The decoded probabilities are also generally low, suggesting that most graph-token hidden states are not strongly aligned with stable vocabulary predictions.

A consistent pattern appears at the sink-token positions g0 and g1 across all three datasets. From approximately layer $20$ onward, these positions frequently decode to \texttt{paper}. This is expected on citation-style datasets to some extent, but it is still informative: TEA-GLM is pretrained on Arxiv and then applied zero-shot to Cora and PubMed, so the repeated appearance of \texttt{paper} suggests that the early graph-token slots retain a broad citation-domain inherited from pretraining. Rather than decoding dataset-specific labels or graph topology information, the sink-token states appear to expose generic domain-level signals.

\xhdr{Summary} This mechanistic evidence is consistent with the intervention results in Sec.~\ref{sec:rq3}. If graph sink tokens were the main carriers of task-critical graph structure, we would expect their decoded states to reflect labels, neighborhood information, or discriminative node semantics. Instead, they mostly decode to generic terms with low confidence. Together, these results suggest that activation-level saliency in graph tokens is not sufficient evidence of graph-semantic utility, and that TEA-GLM does not fully translate its graph-token representations into dataset-specific topology-aware information under zero-shot transfer.

\begin{tcolorbox}
\textbf{Key Takeaway.} Logit lens analysis shows that graph sink tokens mainly expose generic domain-level terms rather than task-specific or topology-aware graph information.
\end{tcolorbox}
\section{Related Work}
\label{sec:related_work}

\xhdr{Graph Language Models}
Graph representation learning has become a central tool for modeling relational data across scientific and real-world domains~\citep{chen2020graph, agarwal2021towards, zhang2026quantifying, besta2024graph, zhang2025efficient, ju2025survey}. Recent work has increasingly explored how LLMs can be adapted to graph-structured data, especially for text-attributed graphs, citation networks, recommendation graphs, and molecular graphs~\citep{jin2024large,li2023survey,ren2024survey,zhang2024text}. Existing approaches can be roughly grouped by the role assigned to the LLM: some use LLMs as predictors over verbalized graph prompts~\citep{ye2024language, chen2024llaga}, some use LLMs as text encoders to enrich node features, and others align graph representations with the LLM embedding space for generative or instruction-following graph tasks~\citep{wang2024llms,zhao2023gimlet}. These methods show that LLMs can provide useful semantic priors for graph learning. 

\looseness=-1\xhdr{Attention Sinks} Attention sinks have been widely observed in Transformer-based models, where certain tokens receive disproportionate attention from other tokens despite limited semantic relevance~\citep{xiao2024duoattention,gu2024attention}. Related work on massive activations shows that sink behavior is often accompanied by extreme hidden-state values on a small number of feature dimensions~\citep{dettmers2022gpt3,sun2024massive, oh2024house}. Recent studies further show that attention sinks are not limited to text-only LLMs, but also appear in vision-language and multimodal models, where visual or perceptual tokens can form sink-like behavior with model- and task-dependent effects~\citep{kang2025see,luo2025sink,zhang2025attention,cappellazzo2025mitigating}.

\section{Conclusion}
\label{sec:conclusion}

We study how GLMs process graph tokens through the lens of graph sink behavior. Across LLaGA and TEA-GLM, we find that graph sink tokens consistently emerge as activation-level outliers, but their internal saliency does not translate into dominant query attention or functional importance for downstream prediction. Our intervention and logit lens analyses further show that graph sink tokens are not the main carriers of graph-semantic or structural information. These results suggest that mapping graph topology into the LLM token space does not by itself guarantee a fully usable topology-aware internal representation.

\looseness=-1\xhdr{Limitations and Future Work} While we have chosen two representative GLM architectures used in current GLM research, other GLM architectures may exhibit different graph sink token behavior. Future work directions include how graph-token construction, placement, and graph-text alignment can be improved so that graph tokens better preserve topology-aware information inside LLMs.

\section{Acknowledgments}
We would like to thank all the anonymous clinicians for their valuable responses to our survey. C.A. and \href{https://chirag-agarwall.github.io/}{Aikyam Lab} is supported, in part, by grants from Capital One, LaCross Institute for Ethical AI in Business, the UVA Environmental Institute, OpenAI Researcher Program, Thinking Machine's Tinker Research Grant, and Cohere. The views expressed are those of the authors and do not reflect the official policy or the position of the funding agencies.

\newpage

\bibliographystyle{unsrtnat}
\bibliography{reference}
\newpage
\appendix
\section{Appendix}
\label{sec:app}


\subsection{More Intervention Results}
\label{app:intervention}

We provide additional intervention results that are omitted from the main paper due to space constraints. Table~\ref{tab:app_rq3_link_intervention_results} reports the link prediction results under pruning and swapping interventions. Table~\ref{tab:rq3_reposition_results} reports the LLaGA repositioning experiment for both node classification and link prediction tasks, where detected graph sink tokens are moved to the front of the graph-token sequence.

The swap and repositioning interventions both alter the original graph-token order and therefore disturb the topology pattern encoded by LLaGA's fixed graph-token template. However, repositioning is a stronger positional intervention than swapping for LLaGA: it moves the detected sink tokens to the front of the graph-token sequence, following the idea of sink-token repositioning as a mitigation technique in prior vision attention sink analysis~\citep{luo2025sink}, but here we use it as a diagnostic intervention. We do not apply repositioning to TEA-GLM because its graph sink tokens already appear at positions $0$ and $1$ in most cases, so moving them to the front would not create a meaningful additional intervention.

\begin{table}[h]
    \centering
    \setlength{\tabcolsep}{4pt}
    \renewcommand{\arraystretch}{0.95}
    \caption{
    Link prediction performance under graph-token interventions.
    We compare the baseline model with pruning the top-$2$ graph sink tokens,
    swapping graph sink and non-sink token positions, and randomly pruning two non-sink graph tokens. \textcolor{red}{Red} values denote cases where non-sink pruning causes a larger performance drop than top-$2$ sink pruning and baseline results.
    }
    \label{tab:app_rq3_link_intervention_results}
    \begin{tabular}{llccc}
        \toprule
        Model & Intervention & Arxiv & Cora & PubMed \\
        \midrule

        \multirow{4}{*}{LLaGA}
        & Baseline & 91.40 & 83.60 & 87.00 \\
        & Top-$2$ Sink & 90.80 & 83.60 & 87.60 \\
        & Non-sink & \textcolor{red}{90.76}\std{0.39} & \textcolor{red}{82.04}\std{0.39} & 87.08\std{0.37} \\
        & Swap & 90.60\std{0.33} & 83.40\std{0.81} & 88.04\std{0.27} \\

        \midrule

        \multirow{4}{*}{TEA-GLM}
        & Baseline & 57.33 & 63.67 & 62.33 \\
        & Top-$2$ Sink & 52.67 & 60.33 & 60.67 \\
        & Non-sink & 54.04\std{3.22} & \textcolor{red}{59.20}\std{2.35} & 61.42\std{2.62} \\
        & Swap & 56.67\std{0.62} & 63.40\std{1.12} & 59.40\std{0.72} \\

        \bottomrule
    \end{tabular}
\end{table}

\begin{table}[h]
    \centering
    \caption{
    Performance comparison for LLaGA under graph sink token repositioning intervention.
    We move the detected graph sink tokens to the front of the graph-token sequence and evaluate the effect on downstream performance.
    }
    \label{tab:rq3_reposition_results}
    \begin{tabular}{llccc}
        \toprule
        Task & Reposition Setting & Arxiv & Cora & Pubmed \\
        \midrule
        \multirow{2}{*}{Node}
        & Top-$2$ Sink to Front & 77.00 & 87.60 & 95.00 \\
        & All Sink to Front & 76.40 & 88.20 & 94.60 \\
        \midrule
        \multirow{2}{*}{Link}
        & Top-$2$ Sink to Front & 91.60 & 82.80 & 87.40 \\
        & All Sink to Front & 89.80 & 82.00 & 85.20 \\
        \bottomrule
    \end{tabular}
\end{table}

\newpage
\subsection{Attention Patterns of Graph Sink Tokens on Link Prediction}
\label{app:rq2_link_attention}
\begin{figure}[h]
    \centering
    \includegraphics[width=\linewidth]{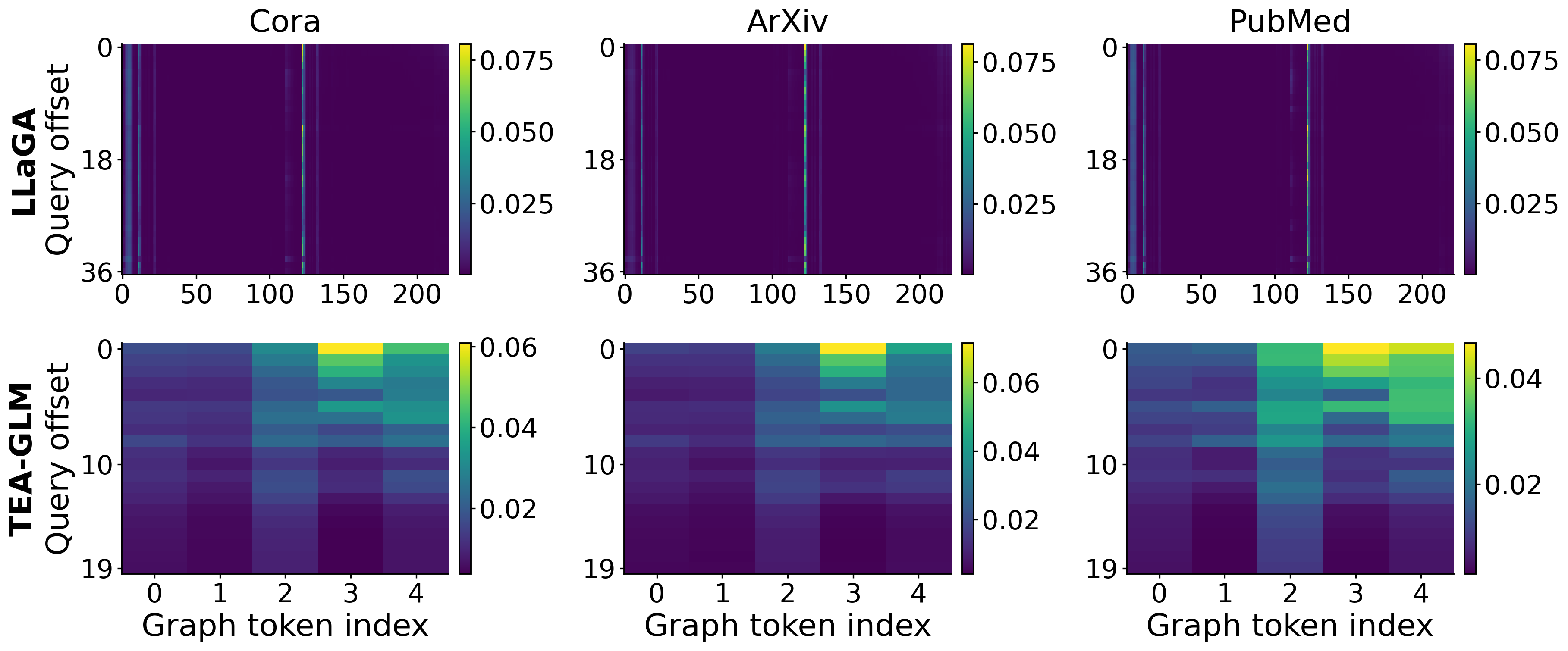}
    \caption{
    \textbf{Query-to-graph attention maps for link prediction.}
    Attention weights are averaged over heads and test samples.
    }
    \label{fig:app_rq2_lp_query_attention}
\end{figure}

\begin{figure}[h]
    \centering
    \includegraphics[width=\linewidth]{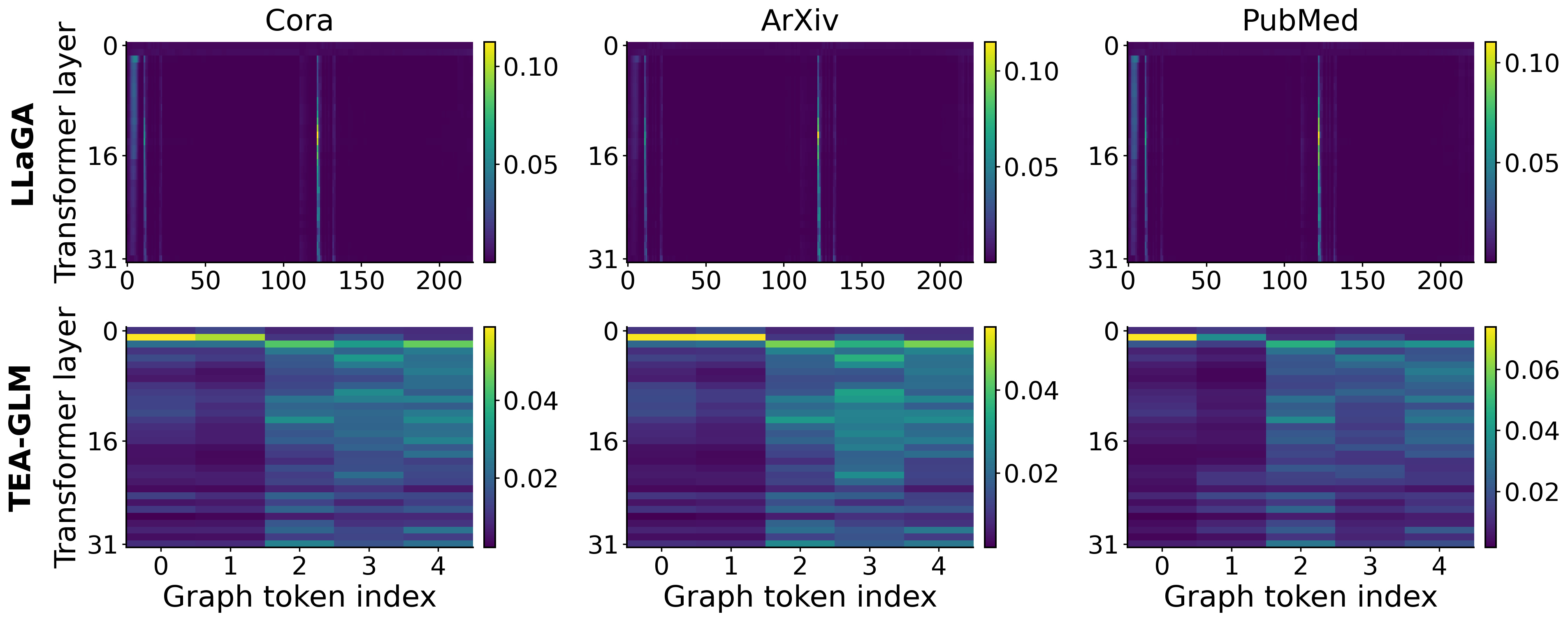}
    \caption{
    \textbf{Layer-wise query-to-graph attention maps for link prediction.}
    Attention weights are averaged over heads and test samples, and the y-axis denotes transformer layers.
    }
    \label{fig:app_rq2_lp_layer_attention}
\end{figure}

\newpage
\subsection{More Logit Lens Results}
\label{app:logit_lens}

\begin{figure}[h]
    \centering
    \includegraphics[width=\linewidth]{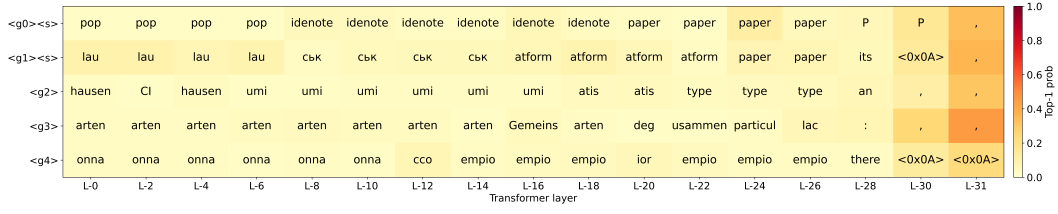}
    \caption{
    \textbf{Logit lens analysis for TEA-GLM graph tokens on Arxiv node classification.}
    Each cell shows the most frequent top-$1$ decoded vocabulary token for a graph-token position and transformer layer; color denotes the averaged probability across selected samples.
    Sink-token positions g0 and g1 frequently decode to generic citation-domain terms such as \texttt{paper} in later layers.
    }
    \label{fig:app_tea_glm_logit_lens_arxiv}
\end{figure}

\begin{figure}[h]
    \centering
    \includegraphics[width=\linewidth]{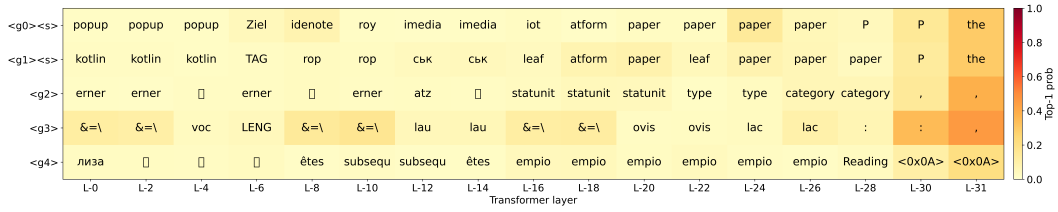}
    \caption{
    \textbf{Logit lens analysis for TEA-GLM graph tokens on Cora node classification.}
    Following the same setup as Fig.~\ref{fig:app_tea_glm_logit_lens_arxiv}, decoded graph-token states mostly correspond to fragmented subwords or generic terms, with later layers showing citation-domain tokens such as \texttt{paper}.
    }
    \label{fig:app_tea_glm_logit_lens_cora}
\end{figure}

\newpage

\newpage
\subsection{Graph Sink Analysis on InstructGLM}
\label{app:instructglm_sink}

\begin{figure}[h]
    \centering
    \includegraphics[width=\linewidth]{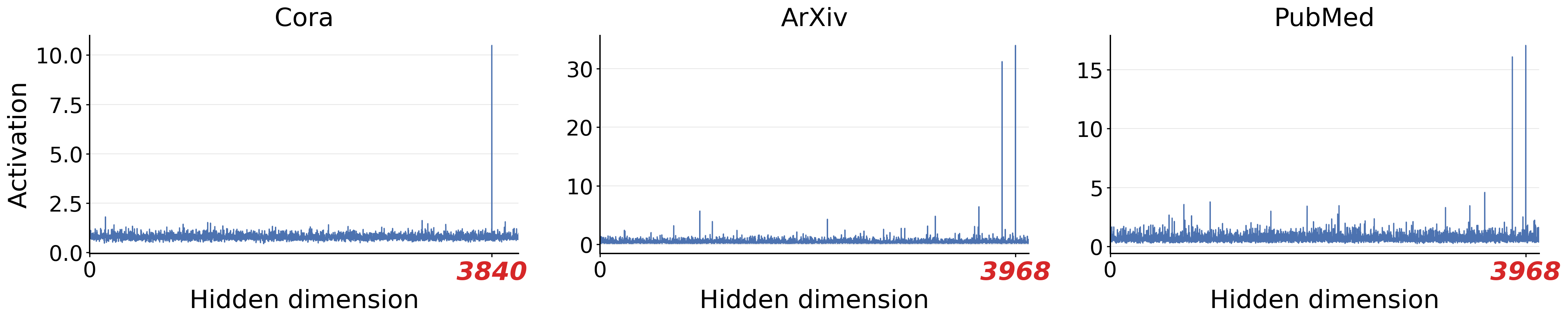}
    \caption{
    \textbf{Activation values across hidden dimensions for detected graph sink tokens in InstructGLM on node classification.}
    Results are averaged over $300$ test samples for Cora, Arxiv, and PubMed.
    InstructGLM also exhibits sparse activation-level outliers, with dominant spikes appearing at a small number of hidden dimensions.
    }
    \label{fig:app_instructglm_sink_dim}
\end{figure}

\begin{figure}[h]
    \centering
    \includegraphics[width=\linewidth]{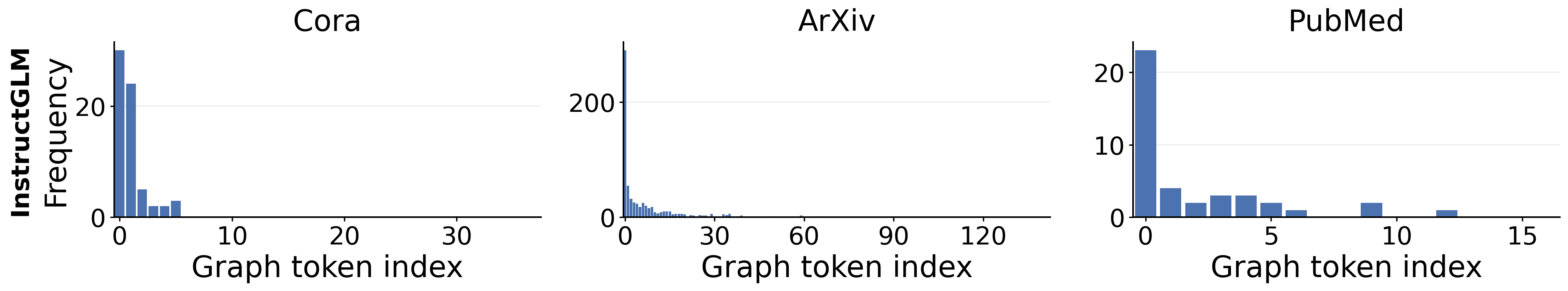}
    \caption{
    \textbf{Distribution of detected graph sink token positions in InstructGLM on node classification.}
    Results are averaged over $300$ test samples for Cora, Arxiv, and PubMed.
    Graph sink tokens are strongly biased toward early graph-token positions across datasets.
    }
    \label{fig:app_instructglm_sink_distribution}
\end{figure}

In addition to LLaGA and TEA-GLM, we conduct a prototyping experiment on sink dimension activation analysis for InstructGLM \citep{ye2024language} on the same three datasets, with $\tau = 10$ for all datasets. Figures~\ref{fig:app_instructglm_sink_dim} and~\ref{fig:app_instructglm_sink_distribution} show that InstructGLM follows the same broad sink pattern observed in the main results. Its detected graph sink tokens are sparse activation-level outliers: most hidden dimensions remain at low magnitude, while only a small number of dimensions produce large spikes. The dominant sink dimension differs across datasets, with dimension $3840$ appearing on Cora and dimension $3968$ appearing on Arxiv and PubMed. 

The position distribution also shows a strong early-token bias. Most graph sink tokens occur near the beginning of the graph-token sequence across all three datasets. These appendix results support the main finding that graph sink tokens emerge as activation-level outliers and tend to concentrate near early graph-token positions across GLM architectures.

\newpage
\subsection{Implementation Details}
\label{app:implementation_details}

\looseness=-1\xhdr{Inference hyperparameters} For both LLaGA and TEA-GLM, all reported results are obtained using deterministic greedy decoding with a fixed random seed of $42$ for reproducibility. Unless otherwise stated, we disable sampling and beam search during generation. All experiments are conducted on two NVIDIA A100 GPUs.

\looseness=-1\xhdr{LLaGA} For LLaGA, we use the publicly released checkpoint with \texttt{vicuna-7b-v1.5-16k} as the language-model backbone. We follow the Vicuna v1 conversation template. Graph nodes are encoded with the SimTeG embedding stack~\citep{duan2023simteg} with a $4096$-dimensional node representation. We use both the Neighborhood Detail (ND) template for node classification and for link prediction, with a sampling fan-out of $10$ neighbors per hop and a maximum hop of $2$. During inference, we call \texttt{model.generate} with \texttt{do\_sample=False}, \texttt{temperature=0.0}, \texttt{top\_p=None}, \texttt{num\_beams=1}, and \texttt{max\_new\_tokens=1024}. We enable KV caching and return hidden states and attention weights for our analyses.

\looseness=-1\xhdr{TEA-GLM}
For TEA-GLM, we train a citation-domain checkpoint on ogbn-Arxiv using \texttt{vicuna-7b-v1.5} as the frozen LLM backbone. During training, only the graph encoder, cross-modal projector, and learned graph-token embeddings are updated. The graph branch is initialized from a pretrained two-layer GraphSAGE encoder with hidden size $2048$ and output dimension $4096$, matching the projector configuration. Each prompt is augmented with $K=5$ learned graph tokens and uses \texttt{max\_text\_length}$=700$ for ogbn-Arxiv. Training is launched with \texttt{accelerate} on two GPUs using a per-device batch size of $8$ and gradient accumulation steps of $2$. We use AdamW with learning rate $1\times10^{-4}$, a cosine learning-rate schedule, and a warmup ratio of $0.03$. The model is trained for $50$ epochs, and we use the final checkpoint for all TEA-GLM transfer evaluations reported in the paper.


\end{document}